\documentclass[runningheads]{llncs}
\usepackage{graphicx}
\usepackage{amsmath}
\usepackage{booktabs}
\usepackage{colortbl}
\usepackage{algpseudocode}
\usepackage{algorithm} 
\usepackage{adjustbox}
\usepackage{color}
\usepackage{amsfonts}
\usepackage{amssymb}
\usepackage{bbding}
\usepackage{multirow}
	
\definecolor{Gray}{gray}{0.9}
\definecolor{LightCyan}{rgb}{0.88,1,1}
\begin{document}
%
\title{Mix-Pooling Strategy for Attention Mechanism\thanks{Work in progress.}}
%
\author{Shanshan Zhong, Wushao Wen\thanks{Corresponding authour}, Jinghui Qin}
%

\institute{School of Computer Science and Engineering \\Sun Yat-sen University}
\maketitle              
\begin{abstract}

Recently many effective attention modules are proposed to boot the model performance by exploiting the internal information of convolutional neural networks in computer vision. In general, many previous works ignore considering the design of the pooling strategy of the attention mechanism since they adopt the global average pooling for granted, which hinders the further improvement of the performance of the attention mechanism. However, we empirically find and verify a phenomenon that the simple linear combination of global max-pooling and global min-pooling can produce pooling strategies that match or exceed the performance of global average pooling. Based on this empirical observation, we propose a simple-yet-effective attention module SPEM, which adopts a self-adaptive pooling strategy based on global max-pooling and global min-pooling and a lightweight module for producing the attention map. The effectiveness of SPEM is demonstrated by extensive experiments on widely-used benchmark datasets and popular attention networks.

\end{abstract}
\section{Introduction}

\begin{figure}[t]
  \centering
  \vspace{-3mm}
  \includegraphics[width=0.8\linewidth]{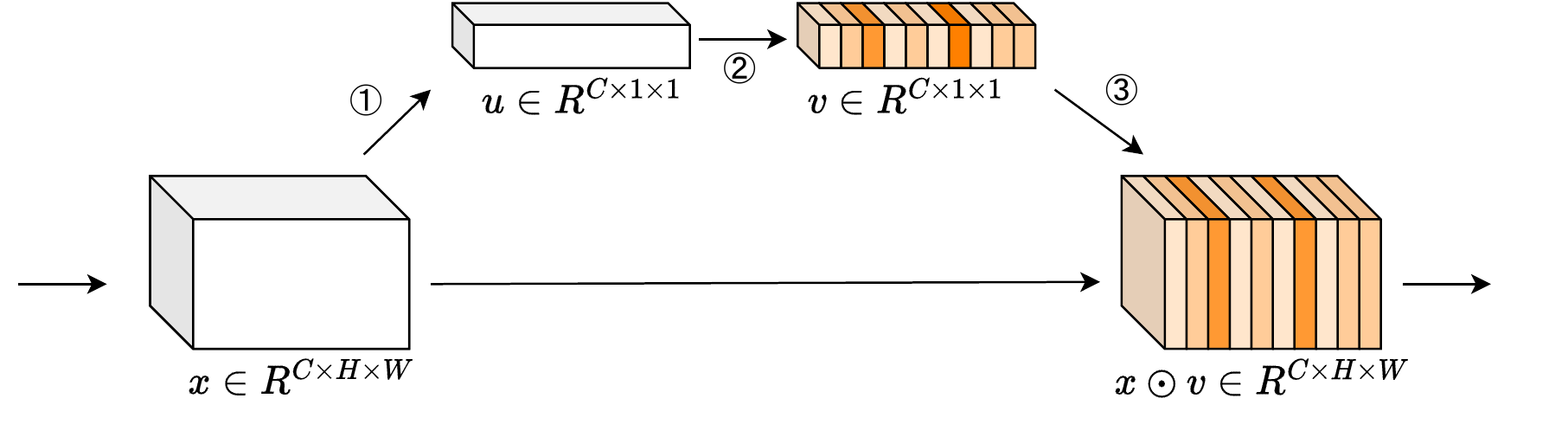}
  \caption{Diagram of each attention sub-module. the attention module can be divided into three stages~\cite{huang2020dianet}: \raisebox{.9pt}{\textcircled{\raisebox{-.9pt}{1}}}pooling, \raisebox{.9pt}{\textcircled{\raisebox{-.9pt}{2}}} excitation; and \raisebox{.9pt}{\textcircled{\raisebox{-.9pt}{3}}} recalibration.}
\label{fig:attention}
\end{figure}

Methods for diverting attention to the most important regions of an image and disregarding irrelevant parts are called attention mechanisms~\cite{guo2022attention}. 
Recently, there has been a lot of interest in considering the use of attention mechanisms in convolutional neural networks~(CNNs) to enhance feature extraction, which has been used for many vision tasks~\cite{fu2019dual,qin2021fcanet,wang2020hierarchical}. 
In these vision tasks, the attention modules are generally elaborate neural network that are plugged into the backbone network. Specifically, most of the existing research is mainly based on three stages~\cite{huang2020dianet}: \raisebox{.9pt}{\textcircled{\raisebox{-.9pt}{1}}} pooling, \raisebox{.9pt}{\textcircled{\raisebox{-.9pt}{2}}} excitation; and \raisebox{.9pt}{\textcircled{\raisebox{-.9pt}{3}}} recalibration. As shown in Fig.~\ref{fig:attention},
in stage \raisebox{.9pt}{\textcircled{\raisebox{-.9pt}{1}}} (pooling), the information of the features map is initially compressed; then in stage \raisebox{.9pt}{\textcircled{\raisebox{-.9pt}{2}}} (excitation), the compressed information is passed through a module to extract the attention maps; finally, in stage \raisebox{.9pt}{\textcircled{\raisebox{-.9pt}{3}}} (recalibration), the attention maps are applied to the hidden layer of the backbone network and adjust the feature. 
For these stages, many previous works~\cite{huang2020dianet,luo2020stochastic,qin2021fcanet} only focus on stage \raisebox{.9pt}{\textcircled{\raisebox{-.9pt}{2}}} (excitation) and stage \raisebox{.9pt}{\textcircled{\raisebox{-.9pt}{3}}} (recalibration), which directly use global average pooling~(GAP) in stage \raisebox{.9pt}{\textcircled{\raisebox{-.9pt}{1}}} to obtain global information embedding for granted due to the simplicity and effectiveness of GAP. 

\begin{figure}
  \centering
  \vspace{-3mm}
  \includegraphics[width=\linewidth]{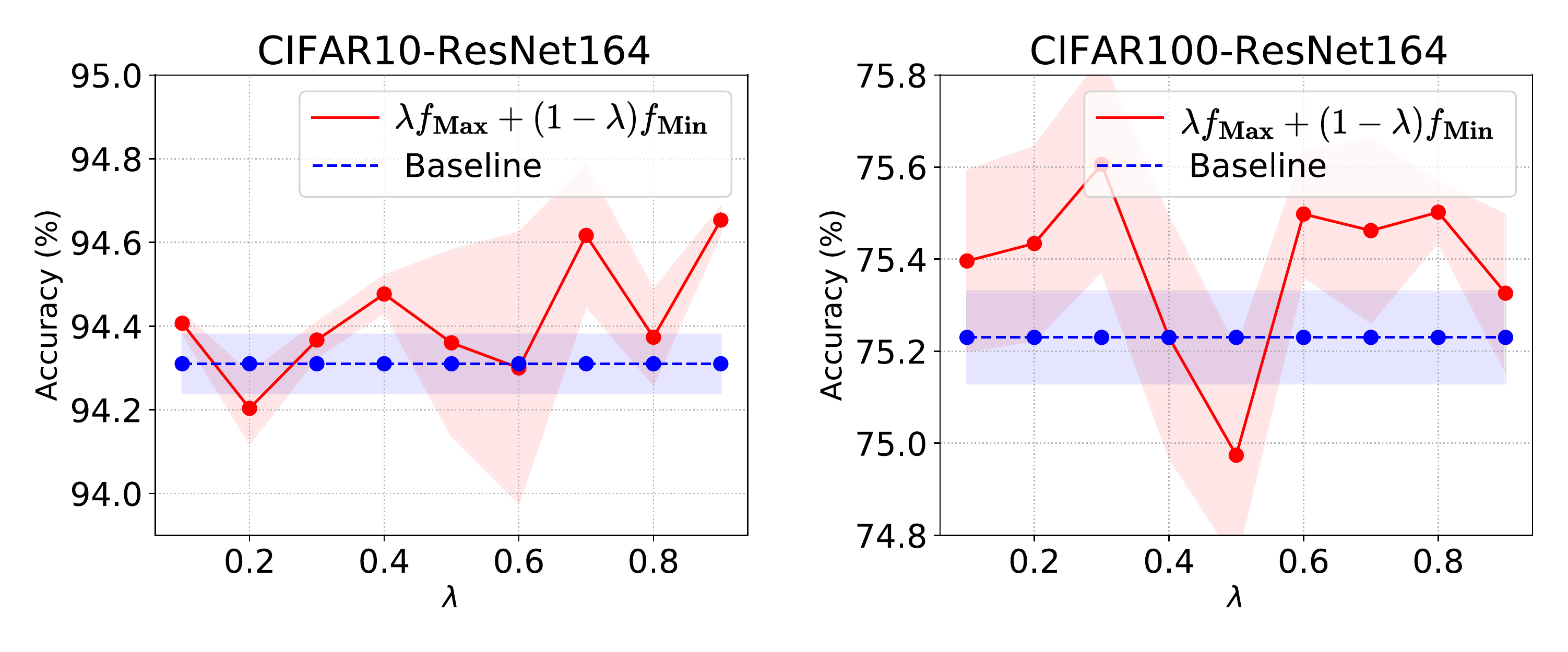}
  \caption{The performance of ResNet164-SPEM with different $\lambda$. As illustrated, both on CIFAR10 and CIFAR100, the performance of different simple linear combination of global max-pooling $f_{\textbf{Max}}$ and global min-pooling $f_{\textbf{Min}}$ can match or exceed the performance of GAP. }
\label{fig:senet}
\end{figure}

However, taking SENet~\cite{hu2018squeeze} with GAP on CIFAR10 and CIFAR100 as baseline, our experiments, as shown in Fig.~\ref{fig:senet}, indicate the optimal pooling strategy is not always GAP, but a simple linear combination of global max-pooling $f_{\textbf{Max}}$ and global min-pooling $f_{\textbf{Min}}$, where, for input $x\in R^{C\times H\times W}$,
\begin{equation}
    \begin{aligned}
&f_{\operatorname{Max}}=[\operatorname{Max}(x[0,:,:]), \ldots, \operatorname{Max}(x[C,:,:])] \in R^{C\times 1 \times 1}, \\
&f_{\operatorname{Min}}=[\operatorname{Min}(x[0,:,:]), \ldots, \operatorname{Min}(x[C,:,:])] \in R^{C\times 1 \times 1}.
\end{aligned}
\end{equation}
Note that, the quality of the attention map depends heavily~\cite{canbek2022gaining,geiger2021garbage,smith1994need} on the quality of the inputs which is transformed based on the pooling strategy. 
Therefore, it is necessary to propose a better pooling strategy than GAP.
In this paper, we propose a simple-yet-effective attention module, SPEM, where we adopt a self-adaptive pooling strategy for constructing global information. Then we design a simple and lightweight attention module for adjusting attention maps. Extensive experiments show that SPEM can achieve the state-of-the-art results without any elaborated model crafting.

The contributions of this paper can be summarized as follows:
\begin{itemize}
    \item We find that the widely-used pooling strategy GAP in the attention mechanism is not an optimal choice for global information construction and can be further improved.
    \item According to the insufficiency of GAP, we propose SPEM with self-adaptive pooling strategy and a lightweight attention module, which can be easily plugged into various neural network architectures.
    \item The extensive experiments of SPEM show that our method can achieve the state-of-the-art results on several popular benchmarks.
\end{itemize}

\section{Related Works}

\noindent\textbf{Convolutional Neural Networks.} For the reason that deep learning can efficiently mine data features, CNNs are widely used in a variety of computer vision tasks~\cite{huang2020efficient,huang2020convolution,he2021blending,huang2021rethinking}. 
VGG~\cite{simonyan2014very}, ResNet~\cite{he2016deep}, and DenseNet~\cite{huang2017densely}are propesed to improve CNNs. 


\noindent\textbf{Attention Mechanism in Vision Recognition.} Mnih et al.~\cite{mnih2014recurrent,2015Show} model the important of features in visual tasks. Subsequently, many works pay attention to attention mechanisms. Ba et al.~\cite{ba2014multiple} propose a deep recurrent neural network trained with reinforcement learning to attend to the most relevant regions of the input image. Wang et al.~\cite{wang2017residual} propose attention residual learning to train very deep Residual Attention Networks which can be easily scaled up to hundreds of layers. 
Yang et al.~\cite{yang2020gated} propose a generally applicable transformation unit for visual recognition with deep convolutional neural networks. Qin et al.~\cite{qin2021fcanet} start from a different view and regard the channel representation problem as a compression process using frequency analysis. 
Besides these works, many reseachers try to extend the attention mechanisms to specific tasks, e.g.  image generation~\cite{gregor2015draw,li2022real}, and super resolution~\cite{qin2020multi,qin2019difficulty,qin2022lightweight}.  

\noindent\textbf{Feature Pooling in Attention Mechanism. }
Since SENet~\cite{hu2018squeeze} uses GAP to get global information embedding, many subsequent studies have used GAP as the way of feature pooling. In addition, some studies have explored the way of feature pooling. Woo et al.~\cite{woo2018cbam} compare GAP and GMP(Global max-pooling) and suggest to use max-pooled features as well.  Luo et al.~\cite{luo2020stochastic} propose Stochastic region pooling to make attention more expressive, which takes GAP to extract the descriptor.
Qin et al.~\cite{qin2021fcanet} analyze the extraction task from the perspective of frequency analysis, which shows that GAP is a special case of frequency domain feature decomposition.

\section{Self-adaptive Pooling Enhanced Attention Module}

\begin{figure}[t]
  \centering
  \vspace{-3mm}
  \includegraphics[width=1\linewidth]{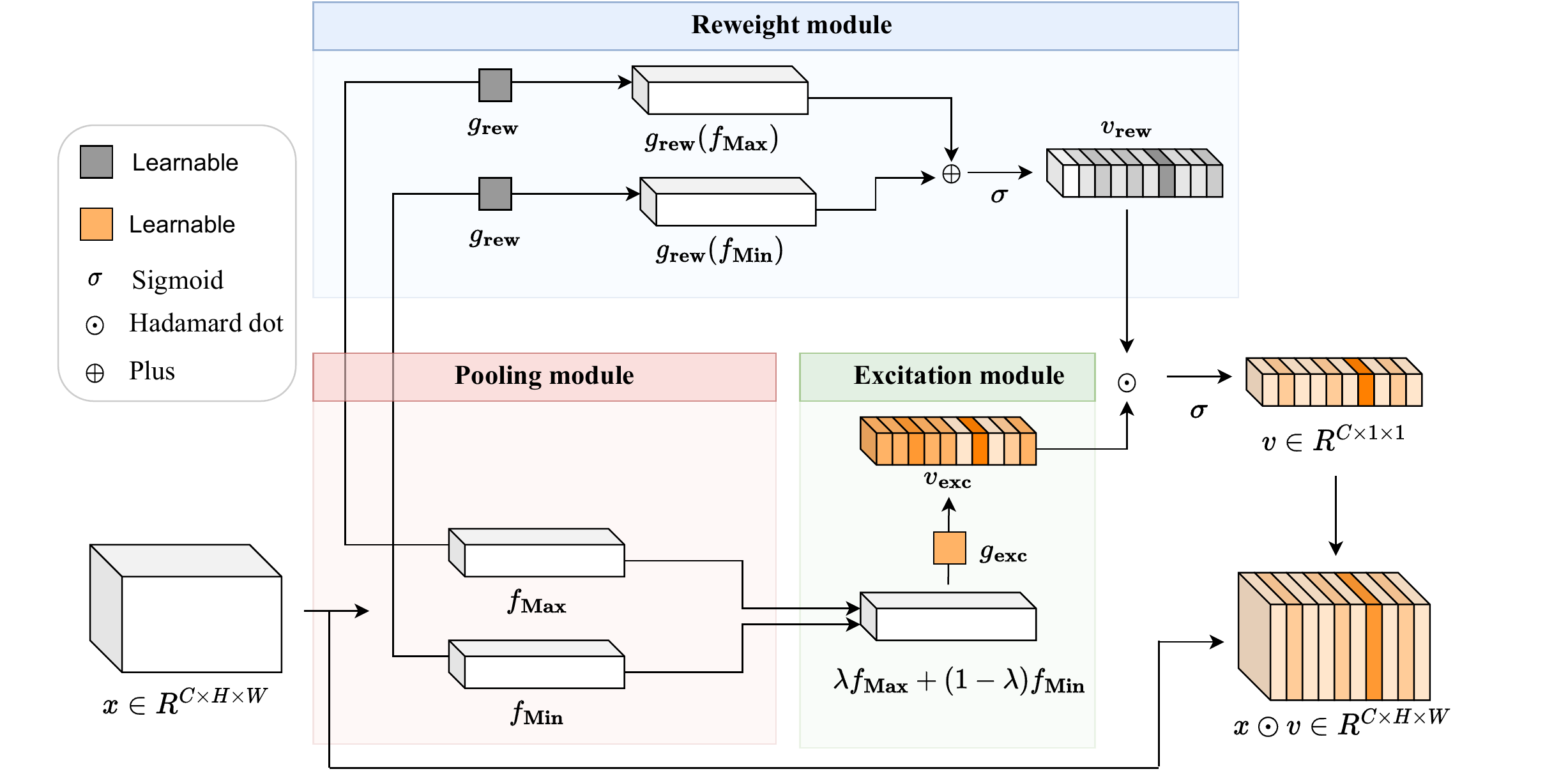}
  \caption{The overall architecture of SPEM.}
\label{fig:arch}
\end{figure}

In this section, we formally introduce SPEM which mainly consists of three modules: pooling module, excitation module, and reweight module. The overall architecture of SPEM  is shown in Fig.~\ref{fig:arch}. Given an intermediate feature map $x \in R^{C \times H \times W}$ as input, we first get a global information embedding $u \in R^{C \times 1 \times 1}$ through a pooling module. Then, $u$ is applied to exciting the attention map $v_{\textbf{exc}} \in R^{C \times 1 \times 1}$ in the excitation module. The reweight module generates the attention map $v_{\textbf{rew}}$ for $v_{\textbf{exc}}$. Finally, we adjust the $v_{\textbf{exc}}$ by $v_{\textbf{rew}}$. The algorithm of our SPEM is shown in Alg.~\ref{alg:ean}. The details of three modules will be introduced detailedly below. 


\subsection{Pooling Module}

In the attention mechanism~\cite{hu2018squeeze}, the efficiency of exploiting channel-wise or spatial-wise contextual information globally depends on how the global information is squeezed. As the experiment results shown in Fig.~\ref{fig:senet}, the pooling strategy GAP is not always optimal for squeezing a global information embedding $u$ since a simple linear combination of the output $f_{\textbf{Max}}$ of a max-pooling operator and the output $f_{\textbf{Min}}$ of a min-pooling operator can achieve higher performance. However, it is laborious and high-cost for us to find suitable $\lambda$ for combining these two operators through experiments. Therefore, we propose a self-adaptive pooling module to generate a global information embedding $u$ by learning a suitable $\lambda$ in a data-driven way.  


Specifically, given an feature map $x$, we can produce a global information embedding $u$ by combining the outputs of the max-pooling operator $f_{\textbf{Max}}$ and the min-pooling operator $f_{\textbf{Min}}$ as follows:
\begin{equation}
u = \lambda f_{\textbf{Max}}(x) + (1-\lambda)  f_{\textbf{Min}}(x),
\label{eq:mix}
\end{equation}
where $\lambda \in [0, 1]$ is a learnable parameter. To make the size of the learnable parameter $\lambda$ belong to the value range $[0, 1]$, we first define two trainable parameters $p_0, p_1$ which are initialized to 0.5. Then we calculate $\lambda$ and $1 - \lambda$ through $p_0$ and $p_1$ as follows:
\begin{equation}
\begin{aligned}
\lambda &= \frac{p_0^2} {p_0^2 + p_1^2}, \\
1 - \lambda &= \frac{p_1^2} {p_0^2 + p_1^2},
\end{aligned}
\label{eq:parameter}
\end{equation}

Here, when $\lambda$ approaches 1 or 0, $u$ will degenerate to $f_{\textbf{Max}}$ or $f_{\textbf{Min}}$, respectively.

\begin{algorithm}[t]  
    \label{alg:ean}   
    \caption{The algorithm of producing attention map from SPEM}
    \textbf{Input:}  An intermediate feature map $x\in R^{C\times H\times W}$; Two learnable transformation $g_{\textbf{exc}}$ and $g_{\textbf{rew}}$; A learnable weight $\lambda$. 
    
    \textbf{Output:} The attention map $v$.
        
    \begin{algorithmic}[1]

    \State\algorithmiccomment{Pooling module}
    
    \State $f_{\textbf{Max}}\gets$ Global max-pooling of $x$
    \State $f_{\textbf{Min}}\gets$ Global min-pooling of $x$
    \State Calculate the Pooling $u \gets \lambda  f_{\textbf{Max}}(x) + (1-\lambda)  f_{\textbf{Min}}(x)$

    
    \State\algorithmiccomment{Excitation module}       
    \State $v_{\textbf{exc}} \gets g_{\textbf{exc}}(u)$
       
    \State\algorithmiccomment{Reweight module} 
        
    \State Calculate $g_{\textbf{rew}}(f_{\textbf{Min}}(x))$ by Eq.~(\ref{eq:detail-rew})
    \State Calculate $g_{\textbf{rew}}(f_{\textbf{Max}}(x))$ by Eq.~(\ref{eq:detail-rew})

    \State $v_{\textbf{rew}} = \sigma ( g_{\textbf{rew}}(f_{\textbf{Min}}(x))  + g_{\textbf{rew}}(f_{\textbf{Max}}(x))) $
    
    \State $v \gets v_{\textbf{exc}} \odot v_{\textbf{rew}}$
    \State\Return $v$
    \end{algorithmic} 
\end{algorithm}  

\subsection{Excitation Module}
To make use of the aggregated global information aggregated to fully capture channel-wise dependencies, we build a lightweight excitation module  $g_{\textbf{exc}} (\cdot)$ to obtain an elementary attention map $v_{\textbf{exc}}$ from the embedding $u$. Our excitation module can be written formally as:
\begin{equation}
v_{\textbf{exc}} = \sigma(g_{\textbf{exc}}(u)) ,
\label{eq:excitation}
\end{equation}
where $g_{\textbf{exc}}(u)$ is designed as a linear transformation $\sigma( \gamma_{\textbf{exc}} \odot u + \beta_{\textbf{exc}})$ inspired by~\cite{liang2020instance}. $\odot$ is dot product. $\gamma_{\textbf{exc}}$ and $ \beta_{\textbf{exc}}$ are two learnable parameters, which are initialized to 0 and -1, respectively. Here, the initialization of $\gamma_{\textbf{exc}}$ and $ \beta_{\textbf{exc}}$ can be referred to \cite{liang2020instance}. After the linear transformation, the sigmoid function $\sigma$ is applied to the value.   


\subsection{Reweighting Module}
To refine the elementary attention map $v_{\textbf{exc}}$ for paying more attention to the most important regions, we propose a reweighting module where the max-pooling operator $f_{\textbf{Max}}$ and the min-pooling operator $f_{\textbf{Min}}$ are applied to fine-tune the attention map $v_{\textbf{exc}}$. The formalization of our reweight module can be defined as follows:
\begin{equation}
v_{\textbf{rew}} = \sigma ( g_{\textbf{rew}}(f_{\textbf{Min}}(x))  + g_{\textbf{rew}}(f_{\textbf{Max}}(x))) ,
\label{eq:reweight}
\end{equation}
where  $v_{\textbf{rew}} \in R^{C \times 1 \times 1}$ is the refined attention map of $v_{\textbf{exc}}$.  $g_{\textbf{rew}}(\cdot)$ is a linear transformation to transform the inputs $f_{\textbf{Min}}(x)$ and $f_{\textbf{Max}}(x)$ into $v_{\textbf{rew}}$. The design of $g_{\textbf{rew}}(\cdot)$ can be defined as follows:
   
\begin{equation}
v_{\textbf{rew}} = \sigma( f_{\textbf{Min}}(x) \odot \gamma_{\textbf{rew}} + f_{\textbf{Max}}(x) \odot \gamma_{\textbf{rew}} + \beta_{\textbf{rew}}),
\label{eq:detail-rew}
\end{equation}
where $\gamma_{\textbf{rew}}$ and $\beta_{\textbf{rew}}$ are initialized to 0 and -1, which are the same as the excitation module.

To obtain final attention map $v$, we use $v_{\textbf{rew}}$ to reweight $v_{\textbf{exc}}$ as follows:
\begin{equation}
v = v_{\textbf{rew}} \odot v_{\textbf{exc}}. 
\label{eq:weight}
\end{equation}
where $\odot$ is dot product.

In this way, the reweighting module is directly applied to optimize the initial attention map $v_{\textbf{exc}}$ via the supplement of global min information and global max information.

\begin{table}[htbp]
  \centering
  \vspace{-3mm}
  \caption{Classification results of different models on CIFAR10 and CIFAR100. }
    \begin{tabular}{lccc}
    \toprule
    \textbf{Model} & \multicolumn{1}{l}{\textbf{Dataset}} & \multicolumn{1}{l}{\textbf{Number of parameters}} & \multicolumn{1}{l}{\textbf{Top-1 acc.}} \\
    \midrule
    ResNet164 & CIFAR10 & 1.70M &   93.39 \\
    ResNet164 + SE~\cite{hu2018squeeze} & CIFAR10 &  1.91M     &  94.24       \\
    ResNet164 + CBAM~\cite{woo2018cbam} & CIFAR10 &  1.92M     &  93.88      \\
    ResNet164 + ECA~\cite{2020ECA} & CIFAR10 &  1.70M     &     94.47    \\
    ResNet164 + SRM~\cite{lee2019srm} & CIFAR10 &  1.74M     &  94.55       \\
    ResNet164 + SGE~\cite{li2019spatial} & CIFAR10 &  1.71M     &    94.25     \\
    \rowcolor{Gray} ResNet164 + SPEM(ours) & CIFAR10 & 1.74M      &  \textbf{94.80} ({\color{red}{$\uparrow$ 1.41}})      \\
    \midrule
    ResNet164 & CIFAR100 &  1.73M     & 74.30       \\
    ResNet164 + SE~\cite{hu2018squeeze} & CIFAR100 & 1.93M      & 75.23     \\
    ResNet164 + CBAM~\cite{woo2018cbam} & CIFAR100 &  1.94M     &  74.68      \\
    ResNet164 + ECA~\cite{2020ECA} & CIFAR100 &  1.72M     &   75.81     \\
    ResNet164 + SRM~\cite{lee2019srm} & CIFAR100 &  1.76M     &   76.01      \\
    ResNet164 + SGE~\cite{li2019spatial} & CIFAR100 & 1.73M      &   74.82     \\
    \rowcolor{Gray} ResNet164 + SPEM(ours) & CIFAR100 &    1.76M   &  \textbf{76.31} ({\color{red}{$\uparrow$ 2.01}})      \\
    \bottomrule
    \end{tabular}%
  \label{tab:main-experiment}%
\end{table}%

\subsection{Loss Function}
Different from the models with other attention modules~\cite{hu2018squeeze,huang2020dianet,2020ECA}, the loss function $L$ for the models with SPEM consists of two parts: the objective function $L_m$ of the major task and the constraint item $p_0^2 + p_1^2$ which is used to prevent $p_0, p_1$ from explosion  during the training, which makes the model training process more stable.

\begin{equation}
L = L_m + \eta (p_0^2 + p_1^2),
\label{eq:loss}
\end{equation}
where $\eta \in R_+$ denotes the penalty coefficient, which is used to control the influence of $(p_0^2 + p_1^2)$ on $L$ and set as 0.1 in our experiments.



\section{Experiments}

In this section, we evaluate the performance of SPEM on image classification and empirically demonstrate its effectiveness.

\begin{table}[htbp]
  \centering
  \vspace{-3mm}
  \caption{Classification results of ResNet-SPEM with different depth on CIFAR10 and CIFAR100. }
    \begin{tabular}{l|ccc}
    \toprule
    \textbf{Model} & \multicolumn{1}{l}{\textbf{Dataset}} & \multicolumn{1}{l}{\textbf{Number of parameters}} & \multicolumn{1}{l}{\textbf{Top-1 acc.}} \\
    \midrule
    ResNet83 & CIFAR10  &   0.87M    &   93.62      \\
    \rowcolor{Gray}ResNet83 + SPEM(ours) & CIFAR10 &  0.89M      &     \textbf{94.03} ({\color{red}{$\uparrow$ 0.41}}) \\
    ResNet164 & CIFAR10 &   1.70M    &   93.39     \\
    \rowcolor{Gray}ResNet164 + SPEM(ours) & CIFAR10 &  1.74M     & \textbf{94.80} ({\color{red}{$\uparrow$ 1.41}})        \\
    ResNet245 & CIFAR10 &   2.54M    &   94.16      \\
    \rowcolor{Gray}ResNet245 + SPEM(ours) & CIFAR10 &  2.59M     &     \textbf{95.08} ({\color{red}{$\uparrow$ 0.92}})   \\
    ResNet326 & CIFAR10 &   3.37M    &   93.45      \\
    \rowcolor{Gray}ResNet326 + SPEM(ours) & CIFAR10 &  3.44M     &     \textbf{95.13 }({\color{red}{$\uparrow$ 1.68}}) \\
    \midrule
    ResNet83 & CIFAR100  &   0.89M    &    73.55     \\
    \rowcolor{Gray}ResNet83 + SPEM(ours) & CIFAR100 &    0.91M    &    \textbf{73.87} ({\color{red}{$\uparrow$ 0.32}})   \\
    ResNet164 & CIFAR100 &   1.73M    &    74.30     \\
    \rowcolor{Gray}ResNet164 + SPEM(ours) & CIFAR100 &   1.76M    &   \textbf{76.31} ({\color{red}{$\uparrow$ 2.01}})      \\
    ResNet245 & CIFAR100 &   2.56M    &    73.88    \\
    \rowcolor{Gray}ResNet245 + SPEM(ours) & CIFAR100 &   2.61M   &     \textbf{76.75} ({\color{red}{$\uparrow$ 2.87}})    \\
    ResNet326 & CIFAR100 &   3.40M    &    72.08    \\
    \rowcolor{Gray}ResNet326 + SPEM(ours) & CIFAR100 &   3.46M   &     \textbf{77.54} ({\color{red}{$\uparrow$ 5.46}})    \\
    \bottomrule
    \end{tabular}%
  \label{tab:depth}%
\end{table}%
\noindent\textbf{Dataset and Implementation Details. } We conduct our experiments on CIFAR10 and CIFAR100~\cite{krizhevsky2009learning}. Both CIFAR10 and CIFAR100 have 50k train images and 10k test images of size 32 by 32 but has 10 and 100 classes respectively. Normalization and standard data augmentation including random cropping and horizontal flipping  are applied to the training data. We deploy SGD optimizer with a momentum of 0.9 and a weight decay of $1e^{-4}$. The total number of training epochs is 164. All models are implemented in PyTorch framework with one Nvidia RTX 2080Ti GPU.

\noindent\textbf{For various self-attention modules.}
We compare our SPEM with other popular self-attention modules based on ResNet164. 
As presented in Table~\ref{tab:main-experiment},
our SPEM achieves higher performance on CIFAR10 and CIFAR100 while it is as lightweight as those existing lightweight self-attention modules, like ECA, SRM and SGE according to the number of parameters. Specifically, comparing to ResNet164, the accuracies are improved by 1.41\% and 2.01\% on CIFAR10 and CIFAR100.


\noindent\textbf{For the depth of networks.} Generally in practice, the ultra-deep neural networks can not guarantee the satisfactory performance due to the vanishing gradient problem \cite{huang2020dianet}. Therefore, we conduct some experiments to show how the depth of backbone network affect the performance of SPEM. 
As shown in Table~\ref{tab:depth}, as the depth increases, our SPEM can mitigate the vanishing gradient to some extent and achieve the significant improvement in very deep network. For example, on CIFAR100, the accuracy of ResNet326 with SPEM can achieve 77.54\% which outperforms ResNet326 over 5.46\%.


\section{Ablation study}
\label{sec:ablation}


In this section, we empirically explore how pooling module and reweight module affect the performance of SPEM.

\begin{table}[htbp]
  \centering
  \vspace{-3mm}
  \caption{Classification results of ablation experiments on the pooling module. }
    \begin{tabular}{ccc}
    \toprule
    \textbf{Combination} & \textbf{Top-1 acc. (CIFAR10)} & \textbf{Top-1 acc.(CIFAR100)}  \\
    \midrule
    $0.1f_{\textbf{Max}} + 0.9f_{\textbf{Min}}$   &  94.32       &  75.83        \\
    $0.3f_{\textbf{Max}} + 0.7f_{\textbf{Min}}$   &  93.96        &  75.55       \\
    $0.5f_{\textbf{Max}} + 0.5f_{\textbf{Min}}$   &  94.67       &  76.11       \\
    $0.7f_{\textbf{Max}} + 0.3f_{\textbf{Min}}$   &  94.72        &  76.03       \\
    $0.9f_{\textbf{Max}} + 0.1f_{\textbf{Min}}$   &  94.33       &  75.88       \\
    \rowcolor{Gray}$\lambda f_{\textbf{Max}} + (1-\lambda)f_{\textbf{Min}}$   &  94.80       &  76.31       \\
    \bottomrule
    \end{tabular}%
  \label{tab:ab-pooling}%
\end{table}%

\noindent\textbf{For the Pooling module. }
In Eq.~(\ref{eq:mix}), the learnable coefficient $\lambda$ is the key part of the pooling strategy in SPEM, which can adjust the proportion of the information between $f_{\textbf{Max}}$ and $f_{\textbf{Min}}$. We investigates the effect of self-adaptive pooling strategy by replacing $\lambda$ with different constants range from 0.1 to 0.9. The experiment results are shown in Table~\ref{tab:ab-pooling}. From the results, we can observe that the performance of self-adaptive parameters is better than  the constant coefficients, which means that self-adaptive pooling strategy can benefit the training of deep models with SPEM. 

Besides, it must be noticed that the same constant coefficient may not get the consistent performance on different datasets.$0.5f_{\textbf{Max}}+0.5f_{\textbf{Min}}$ gets the best result on CIFAR100 while $0.7f_{\textbf{Max}}+0.3f_{\textbf{Min}}$ is the best constant coefficient on CIFAR10. On the contrary, our self-adaptive pooling strategy can achieve better performance on both CIFAR10 and CIFAR100. Therefore, the self-adaptive pooling strategy is suitable for different tasks and datasets.



\begin{table}[htbp]
  \centering
  \vspace{-3mm}
  \caption{Classification results of ablation experiments on the reweighting module. Module frameworks corresponding to different serial numbers are shown in Fig.~\ref{fig:ablation}. "Without RM" means removing the reweight module from SPEM. }
    \begin{tabular}{cccc}
    \toprule
    \textbf{Reweight Module~(RM)} & \multicolumn{1}{l}{\textbf{Dataset}} & \multicolumn{1}{l}{\textbf{Number of parameters}} & \multicolumn{1}{l}{\textbf{Top-1 acc.}}  \\
    \midrule
    (a)   & CIFAR10 &  1.75M     &    94.77     \\
    (b)   & CIFAR10 & 1.74M       &    92.98     \\
    (c)   & CIFAR10 &  1.74M      &    94.43    \\
    (d)   & CIFAR10 &  1.74M      &     93.68    \\
    (e)   & CIFAR10 &  1.74M      &      94.32   \\
    (f)   & CIFAR10 &  1.74M      &      93.98   \\
    (g)   & CIFAR10 &  1.74M      &      94.16   \\
    Without RM  & CIFAR10 &  1.72M     &   93.87      \\
    \rowcolor{Gray}Ours & CIFAR10 & 1.74M & \textbf{94.80}\\
    \midrule
    (a)   & CIFAR100 & 1.78M      &   76.11      \\
    (b)   & CIFAR100 & 1.76M      &     74.21    \\
    (c)   & CIFAR100 & 1.76M      &     75.51    \\
    (d)   & CIFAR100 & 1.76M      &     75.48    \\
    (e)   & CIFAR100 & 1.76M      &     76.11    \\
    (f)   & CIFAR100 & 1.76M      &     75.38    \\
    (g)   & CIFAR100 & 1.76M      &     76.14    \\
    Without RM  & CIFAR100 &  1.74M     &     75.09    \\
    \rowcolor{Gray}Ours & CIFAR100 & 1.76M & \textbf{76.31}\\
    \bottomrule
    \end{tabular}%
  \label{tab:ab-reweight}%
\end{table}%

\noindent\textbf{For the reweight module. } We conduct eight different experiments for the reweight module to explore how the reweight module affect the performance of SPEM, which can be mainly split into four parts.

\begin{figure}[t]
  \centering
  \vspace{-3mm}
  \includegraphics[width=0.8\linewidth]{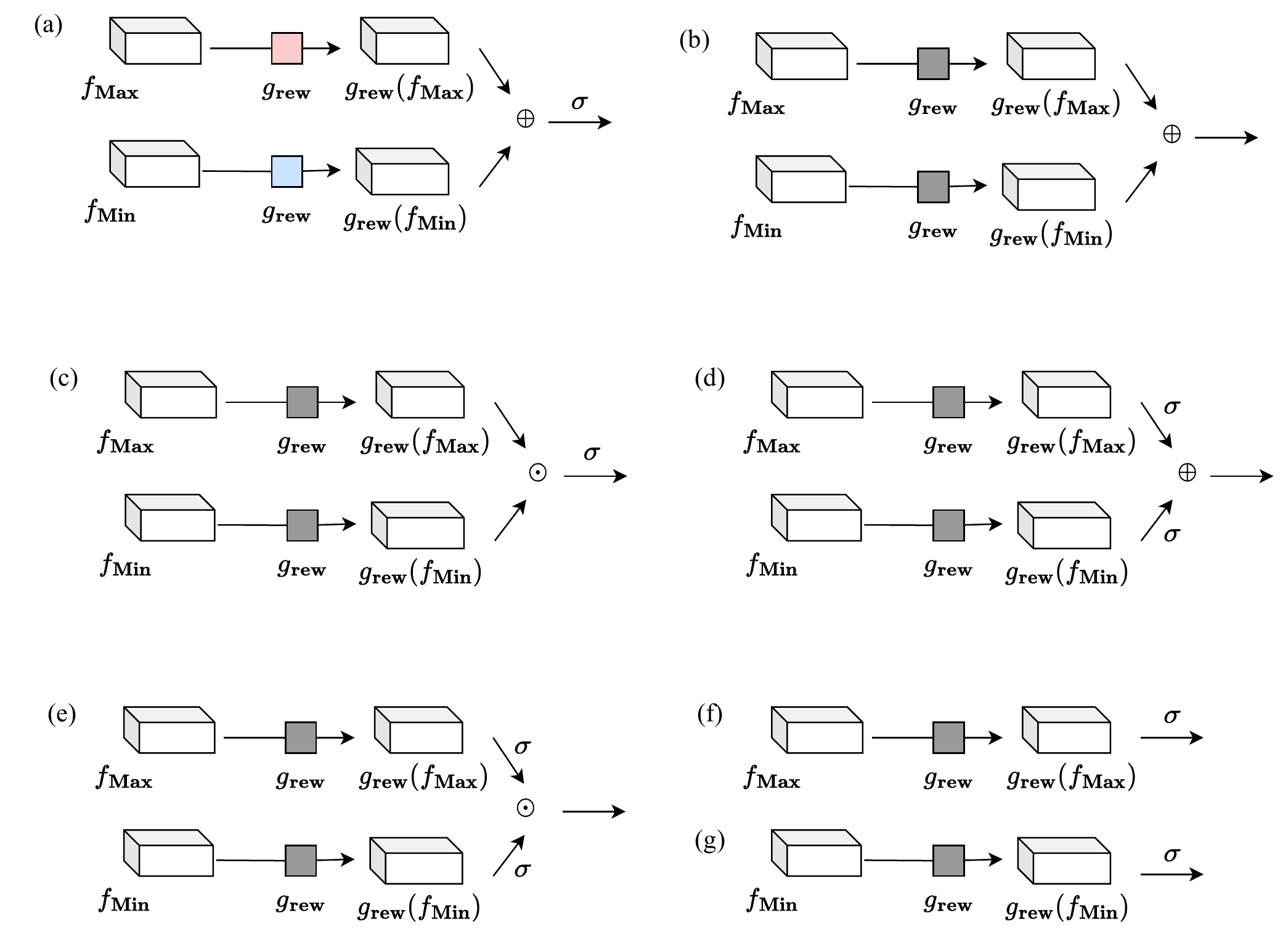}
  \caption{Different frameworks of the reweighting module in ablation experiments. Table \ref{tab:ab-reweight} is the experiment results of corresponding frameworks. (a): $g_{\textbf{rew}}$ is not shared by $f_{\textbf{Min}}(x)$ and $f_{\textbf{Max}}(x)$. (b): calculating $v_{\textbf{rew}}$ without $\sigma$. (c): using $\odot$ to combine $g_{\textbf{rew}}(f_{\textbf{Max}})$ and $g_{\textbf{rew}}(f_{\textbf{Min}})$. (d): using $\sigma$ before adding $g_{\textbf{rew}}(f_{\textbf{Max}})$ and $g_{\textbf{rew}}(f_{\textbf{Min}})$. (e)using $\sigma$ before adding $g_{\textbf{rew}}(f_{\textbf{Max}})$ and $g_{\textbf{rew}}(f_{\textbf{Min}})$ and using $\odot$ to combine $g_{\textbf{rew}}(f_{\textbf{Max}})$ and $g_{\textbf{rew}}(f_{\textbf{Min}})$. (f): only using $f_{\textbf{Max}}$ in the reweight module. (g): only using $f_{\textbf{Min}}$ in the reweighting module. }
\label{fig:ablation}
\end{figure}

First, we experimentally verify that sharing $g_{\textbf{rew}}$ can enable satisfactory performance. The framework of reweighting module based on unshared $g_{\textbf{rew}}$ is shown in Fig.~\ref{fig:ablation}(a), whose accuracy is shown in Table~\ref{tab:ab-reweight}. It shows that shared $g_{\textbf{rew}}$ with less learnable parameters is better than unsharing $g_{\textbf{rew}}$, which indicates that sharing $g_{\textbf{rew}}$ is enough to extract global pooling information well. As a brief conclusion, we use sharing $g_{\textbf{rew}}$ in our SPEM.

Second, we explore Eq.~(\ref{eq:detail-rew}) which combines $g_{\textbf{rew}}(f_{\textbf{Max}})$ and $g_{\textbf{rew}} (f_{\textbf{Min}})$. We construct four combinations as shown in Fig.~\ref{fig:ablation} (b), (c), (d), (e). As shown in Table~\ref{tab:ab-reweight}, we find that the performance of (b) is poor. (d) increases computational complexity but gets a uncompetitive result. (c) is better than (e) on CIFAR10 while (e) is better than (c) on CIFAR100, showing that $\odot$ performs unsteadily. In summary, our stable and reliable SPEM outperforms all of combinations.

Third, we analyze the effects of $f_{\textbf{Max}}$ and $f_{\textbf{Min}}$ respectively. Fig.~\ref{fig:ablation}(f) and Fig.~\ref{fig:ablation}(g) show the frameworks of the experiment settings. By comparing the result of (f) and (g), it can be see that $f_{\textbf{Max}}$ is more effective than $f_{\textbf{Min}}$, which is consistent with the general cognition that the part of a picture with small pixel values is often the ignored part without useful information~\cite{he2010single}. However, although the accuracy of (f) is more than that of (g), combing (f) and (g) can improve the performance of SPEM, which indicates that both $f_{\textbf{Max}}$ and $f_{\textbf{Min}}$ are crucial.

Finally, we further perform an experiment by removing the reweight module from SPEM to explore whether the reweighting module is redundant. we can clearly see that the reweighting module can improve the accuracy of SPEM, showing the importance of this module.

In ablation studies, we mainly explore effects of thepooling module and reweight module. On one hand, we verify that the self-adaptive pooling strategy is more efficient than the constant coefficients. On the other hand, we conduct a series of exploratory experiments on the reweighting module, and the results show that sharing $g_{\textbf{rew}}$ and applying $\sigma$ to activate the sum of $g_{\textbf{rew}}(f_{\textbf{Max}})$ and $g_{\textbf{rew}}(f_{\textbf{Min}})$ are simple-yet-effective.

\section{Conclusion and Future Works}

In this paper, we study the self-adaptive feature pooling strategy for attention mechanism on image recognition and present an self-attention pooling enhanced attention module SPEM to improve representation power of CNN networks. Specifically, we find that a simple linear combination of global max-pooling and global min-pooling is a superior pooling strategy. Therefore, we design a self-adaptive pooling module to replace GAP. Then in consideration of module efficiency, we build a simple and lightweight excitation module to obtain the attention map. Besides, we make full use of global pooling information and use a reweight module to refine the attention map. We further demonstrate empirically the effectiveness of SPEM on two benchmark datasets. The results show that our SPEM is more effective than existing attention modules. 


\bibliographystyle{splncs04}
\bibliography{ref} 
\end{document}